\def\year{2019}\relax
\documentclass[letterpaper]{article} 
\usepackage{aaai19}  
\usepackage{times}  
\usepackage{helvet}  
\usepackage{courier}  
\usepackage{url}  
\usepackage{graphicx}  
\usepackage{amsmath,amssymb}
\usepackage[usenames, dvipsnames]{color}

\begin{document}
%
\title{ReDecode Framework for Iterative Improvement in Paraphrase Generation}

\author{
Milan Aggarwal \\ milaggar@adobe.com \And Nupur Kumari \\ nupkumar@adobe.com \And Ayush Bansal \\ abansal1008@gmail.com \And Balaji Krishnamurthy \\ kbalaji@adobe.com
}

 \maketitle
\begin{abstract}
Generating paraphrases, that is, different variations of a sentence conveying the same meaning, is an important yet challenging task in NLP. Automatically generating paraphrases has its utility in many NLP tasks like question answering, information retrieval, conversational systems to name a few. In this paper, we introduce iterative refinement of generated paraphrases within VAE based generation framework. Current sequence generation models lack the capability to (1) make improvements once the sentence is generated; (2) rectify errors made while decoding. We propose a technique to iteratively refine the output using multiple decoders, each one attending on the output sentence generated by the previous decoder. We improve current state of the art results significantly - with over $9\%$ and $28\%$ absolute increase in \textit{METEOR} scores on $Quora\; question\; pairs$ and $MSCOCO$ datasets respectively. We also show qualitatively through examples that our re-decoding approach generates better paraphrases compared to a single decoder by rectifying errors and making improvements in paraphrase structure, inducing variations and introducing new but semantically coherent information.

\end{abstract}

\section{Introduction}
Paraphrases refer to texts which express the same meaning in different ways. For example, \textit{"Can time travel ever be possible?"} and \textit{"Is time travel a possibility?"} are paraphrases of each other. Human conversations typically involve a high level of paraphrasing to express similar intent, but comprehending such sentences as semantically similar and generating them is a difficult task for a machine. Automatic paraphrase generation is an important task in NLP that has practical significance in many text-to-text generation tasks such as question answering, conversational systems, information retrieval, summarization, etc. Knowledge-based QA systems are highly sensitive to the way a question is asked. Using paraphrases of the asked question while ranking answers in the knowledge base improves the system performance \cite{dong2017learning}. Paraphrasing also fosters incorporating variations in domain specific conversational bots which have a fixed set of responses to prevent them from being repetitive. In the task of query reformulation, paraphrasing has direct utility, e.g. in search engines, paraphrase generation module can be used for recommending different possible variations of the user query or directly show the search results after incorporating the variations as part of the search process. In the case of end-to-end conversational systems, training data can be augmented with paraphrases of available dialogues which helps in improving semantic understanding capability of the system. 

Early paraphrase generation systems used handcrafted rule-based systems \cite{mckeown1983paraphrasing}, relied on automatic extraction of paraphrase patterns from available parallel corpus data \cite{barzilay2003learning} or used knowledge base like word-net for paraphrase generation \cite{bolshakov2004synonymous}. Statistical machine translation tools have also been applied for paraphrase generation \cite{monolingual-machine-translation-for-paraphrase-generation}. These approaches are limited because of their methodology and don't generalize well.

Recent advances in deep neural network based models for sequence generation has advanced state of the art in various NLP tasks such as machine translation \cite{bahdanau2014neural} and question answering \cite{yin2015neural}. For the task of paraphrase generation, \citeauthor{prakash2016neural} \shortcite{prakash2016neural} for the first time explored sequence-to-sequence (Seq2Seq) \cite{sutskever2014sequence} based neural network model and proposed an improved variant of the model - a stacked LSTM Seq2Seq network with residual connections.

In this paper, we present a framework for automatic paraphrase generation which is based on variational autoencoder (VAE)\cite{kingma2013auto}. VAE is used extensively for generative tasks in image domain and has been experimented with in text domain \cite{bowman2015generating} as well; the model usually consists of LSTM RNN \cite{sundermeyer2012lstm} as encoder and decoder (VAE-LSTM), for processing sequential input. Unlike the traditional reconstruction task of VAE, paraphrasing involves generating outputs which are different in their expression but have same semantic meaning. To achieve this objective, \citeauthor{gupta2017deep} \shortcite{gupta2017deep} introduced a supervised variant (VAE-S) of VAE-LSTM where decoder is conditioned on the vector representation of input sentence obtained through another RNN, instead of only depending on latent representation. Our approach is based on the supervised generative sequence modeling through VAE where supervision is obtained through decoder attending over the hidden states of an LSTM RNN that encodes the input sentence.

In this work, we introduce a methodology for iterative improvement of output using the VAE-S framework as compared to previous sequence generation models that decode output sequence only once. This concept is inspired from the idea that given a crude paraphrase and the original sentence, the model should be able to generate a better quality paraphrase in the next iteration by rectifying errors and identifying regions of improvement; similar to what humans can do. We achieve the task of iterative improvement by having multiple decoders in the model and each decoder, except the first one, attends on the output of previous decoder for supervision. We establish the effectiveness of this approach for the domain of paraphrase generation by showing significant improvements in the scores of standard metrics on benchmark paraphrase datasets over state of the art. Our approach is applicable to any other domain which involves sequence generation such as conversational systems, question answering etc. However we do not explore its capabilities in other domains in this paper. Our contributions in this paper can be listed as:

\begin{itemize}
\item We introduce an iterative improvement framework for the output using multiple decoders under VAE based generative model. The first decoder is conditioned on the input sentence encoding whereas further decoders are conditioned on the outputs generated by preceding decoders.
\item We improve the existing state of the art in paraphrase generation task by a significant margin using our above mentioned approach.  
\end{itemize}

\section{Related Work}
Paraphrase generation has been modeled as a Seq2Seq learning problem from the input sentence to the target paraphrase. The first Seq2Seq neural network based approach for this was proposed by \citeauthor{prakash2016neural} \shortcite{prakash2016neural} which was a stacked LSTM RNN model with residual connections. The authors compared it with other variants of Seq2Seq model which included attention and bidirectional LSTM unit. \citeauthor{cao2017joint} \shortcite{cao2017joint} introduced a Seq2Seq model fusing two decoders, one of them is copying decoder and the other is a restricted generative decoder inspired from the human way of paraphrasing that principally involves copying or rewriting. \citeauthor{gupta2017deep} \shortcite{gupta2017deep} introduced VAE  based model for paraphrase generation. VAE as introduced by \citeauthor{kingma2013auto} \shortcite{kingma2013auto} is a generative deep neural network model that maps the input to latent variables and decodes the latent variable to reconstruct the data. VAE is ideal for generating new data as it explicitly learns a probability distribution on the latent code from which a sample is used for decoding. \citeauthor{gupta2017deep} \shortcite{gupta2017deep} condition the decoder on input sentence and use reference paraphrase along with input sentence as input for generating latent code to obtain better quality paraphrases. 

There has also been some work on improving paraphrase generation models inspired from machine translation. It has been shown that paraphrase pairs obtained using back-translated texts from bilingual machine translation corpora has data quality at par with manually-written English paraphrase pairs \cite{wieting2017learning}. There has been work done in syntactically controlled paraphrase generation as well where parse tree template of paraphrase to be generated is also given as input \cite{iyyer2018adversarial}.

Our work in paraphrase generation is similar to the approach of \citeauthor{gupta2017deep} \shortcite{gupta2017deep} in that our model is also based on VAE. The main difference lies in our methodology to iteratively improve the decoded output and use attention mechanism to condition the decoder on input sentence while training. We also introduce a specific loss term to promote generation of varied paraphrases of a given sentence.



\begin{figure*}[t]
\begin{center}
\includegraphics[width=\textwidth,height=3.4in]{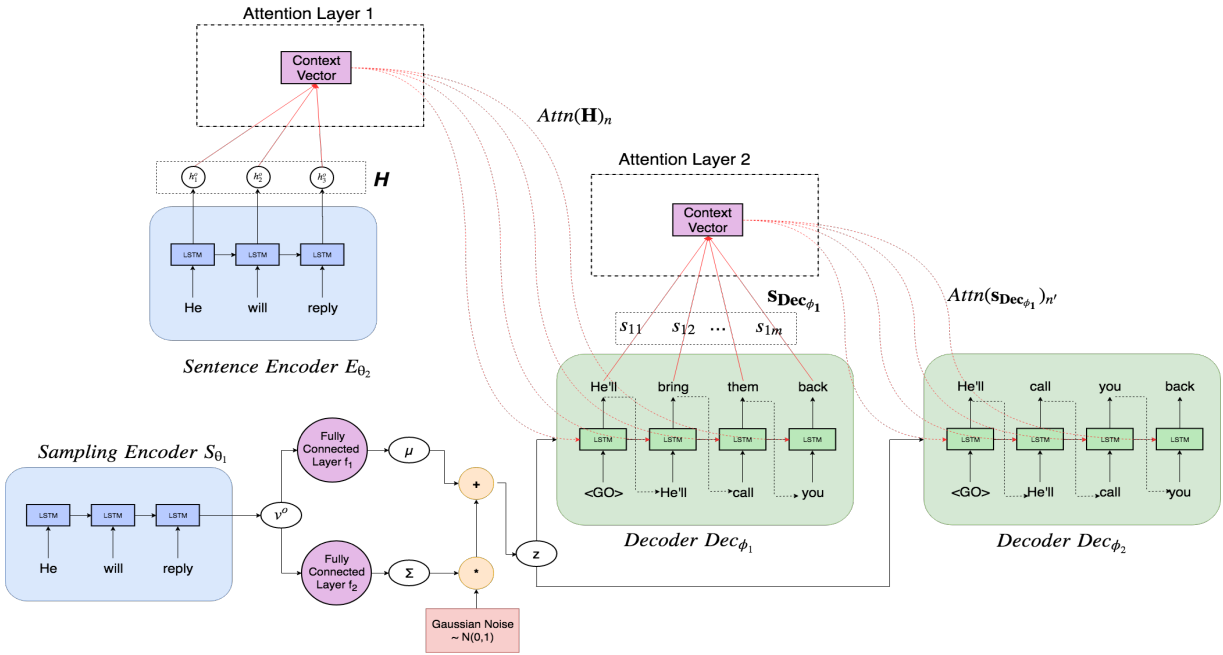}
\caption{Architecture diagram of iterative approach for the case of two decoders $(i=2)$. $S_{\theta_{1}}$ processes the input sentence and its final output is used to obtain the mean and variance vectors $\mu$ and $\Sigma$ through fully connected layer $f_1$ and $f_2$ respectively which are used to sample $z$. $E_{\theta_{2}}$ produces output vectors $\{h_{oi}^d\}$ as it processes the input sentence word by word. First decoder $Dec_{\phi_{1}}$ attends on these output vectors, takes $z$ as input and outputs the paraphrase using standard Seq2Seq technique. Decoder $Dec_{\phi_2}$ attends on the softmax vectors $\{s_{1j}\}$ produced by $Dec_{\phi_1}$ and generates the final output. The dotted connections in the decoders across different time steps show that output generated at time t is passed as input at time t+1 during inference while inputs are pre-determined during training as per teacher forcing technique \cite{williams1989learning}.} 
\label{fig:architecture_image}
\end{center}
\end{figure*}

\section{Methodology}
In this section, we explain our model architecture, which is based on VAE. We first give a brief overview of VAE and then explain our framework in detail.

\subsection{Variational Autoencoder}
Variational Autoencoder as introduced by \citeauthor{kingma2013auto} \shortcite{kingma2013auto} is a generative model that learns a posterior distribution over latent variables for generating output. Input data \textit{x} is mapped to a latent code \textit{z} from which \textit{x} can be reconstructed back. It differs from traditional auto-encoders in the sense that instead of learning a deterministic mapping function to latent code $z=f(x)$, it learns a posterior distribution $p_{\phi}(z|x)$ from the data starting with a prior $p_{\theta}(z)$. The posterior distribution is usually taken to be $N(\mu , \Sigma)$ and prior distribution as $N(0,I)$ to facilitate stochastic back propagation based training. The encoder can be a neural network with a feed forward layer at the end to estimate $\mu$ and $\Sigma$ from $x$. $z$ is sampled from the normal distribution $N(\mu , \Sigma)$ and passed to the decoder as input. The decoder which is also a neural network learns the probability distribution $p_{\theta}(x|z)$ to reconstruct input data from latent code. The network is trained by maximizing the following objective function:
\begin{equation} \label{eqone}
	L(x) =  E_{p_{\phi}(z|x)}[log\;p_{\theta}(x|z)] - D_{KL}(q_{\phi}(z|x) || q_{\theta}(z))
\end{equation}
$\phi$ and $\theta$ are the parameters of encoder and decoder respectively; $D_{KL}$ stands for KL divergence. The objective function maximizes the log likelihood of reconstructed data from the posterior and at the same time reduces the KL divergence between the prior and posterior distribution of latent code $z$. This objective is a valid lower bound on the true log likelihood of data as shown by the authors, therefore maximizing it ensures that the total log likelihood of data is maximized. The first term in equation \ref{eqone} is maximized by minimizing the cross entropy error over the training dataset. 

Since VAE learns the probability distribution $p_{\theta}(x|z)$, it is ideal for generative modeling tasks. For sequence generation task in text, \citeauthor{bowman2015generating} \shortcite{bowman2015generating} proposed RNN based variational autoencoder model. Both the encoder and decoder are LSTM RNN with a feed forward layer at the end of encoder to estimate $\mu$ and $\Sigma$. They introduce techniques like KL cost annealing and word dropout in decoder for efficient learning. \citeauthor{gupta2017deep} \shortcite{gupta2017deep} improved upon this model in a supervised setting of paraphrase generation by conditioning decoder on input sentence encoding computed by a separate encoder and using $z$ at every time step of decoding as input. From now on we use VAE-S (S stands for supervision) to denote this model.  

In our model as well, $z$ is concatenated with the word encoding as input to decoder (as in standard Seq2Seq technique) and decoder is conditioned on input sentence. Since paraphrase generation is subtly different from sentence reconstruction, using $z$ alone may not result in good paraphrases. We condition the decoder on the inputs using well known \textit{Attention Mechanism} \cite{luong2015effective} while generating the paraphrases to enable the model to learn phrase level semantics. \textit{Attention Mechanism} has been widely used in sequence tasks such as \textit{Recognizing Text Entailment} (RTE) \cite{rocktaschel2015reasoning}, \textit{Machine Translation} (MT) \cite{vaswani2017attention} etc.

We explain our attention based ReDecode model architecture in the next section.

\subsection{Model Architecture}
Training data consists of sentence $x_o:\{x_0^o,x_1^o,...x_n^o\}$ and its expected paraphrase $x_p:\{x_0^p,x_1^p,...x_m^p\}$. Input to the model is a sequence of vector encodings of $\{x_i^o\}$ represented as $\{e_i^o\}$ which we take as pre-trained Glove \cite{pennington2014glove} vector embeddings instead of training word vectors from the scratch. The architecture diagram of our model is as shown in figure \ref{fig:architecture_image}. It consists of a Sampling Encoder ($S_{\theta_1}$), Sentence Encoder ($E_{\theta_2}$) and sequence of decoders $\{Dec_{\phi_i}\}$; $\theta_1$, $\theta_2$ and $\phi_i$ are parameters of the model respectively. 
Below we explain each module and training strategy in detail. 

\subsubsection{Sampling Encoder}: $S_{\theta_1}$ is used to encode the original sentence for sampling the latent vector $z$. As shown in figure \ref{fig:architecture_image}, it consists of a single layer LSTM RNN that sequentially processes the word embeddings $\{e_i^o\}$ of words $\{x_i^o\}$ in the original sentence and creates the vector representation $v^o$ of sentence. $v_o$ is then passed through two separate fully connected layers $f_1$ and $f_2$ to estimate mean ($\mu$) and variance($\Sigma$). Final latent code $z$ is sampled from $N(\mu , \Sigma)$ distribution. 

\subsubsection{Sentence Encoder}: $E_{\theta_2}$ computes a vector representation of the input sentence used for generating the output paraphrase in the decoding stage. It is a two layer stacked LSTM unit which sequentially processes the input sentence and generates a set of hidden vectors $H = {h_{1}^o,h_{2}^o, ..., h_{n}^o}$ corresponding to each time step of the input sequence. These hidden vectors are attended upon by the decoder. In attention mechanism, given a sequence of vectors $\{m_i\}$, attributed as memory \textbf{M} with vectors arranged along the columns, the decoder LSTM learns a context vector derived using weighted sum of columns of \textbf{M} as a function of its input and hidden state at time step j and uses it for generating the output. The decoder learns to identify and focus on specific parts of the memory while generating the words.


\subsubsection{Iterative Decoder} : In the decoding stage, we propose to use multiple decoders $Dec_{\phi_1}$, $Dec_{\phi_2}$, ..., $Dec_{\phi_n}$ to generate the output iteratively. While training, the input to each decoder is $z$ sampled using $S_{\theta_1}$ concatenated with encoding $e_i^p$ of $x_i^p$ at each time step. During inference, the generated word is given as input to the next step of decoding as in standard Seq2Seq paradigm. Each decoder is a two layer stacked LSTM unit followed by a projection layer which outputs a likelihood distribution over the vocabulary. In addition, decoder $Dec_{\phi_i}$ attends on the softmax vectors generated by $Dec_{\phi_{i-1}}$ whereas $Dec_{\phi_1}$ attends over the outputs $\{h_{i}^o\}$ generated by Sentence Encoder $E_{\theta_2}$. More formally we iteratively generate a sequence of paraphrases $\{\mathbf{p_1},\mathbf{p_2}, ..., \mathbf{p_n}\}$ such that,

\begin{equation} \label{eqthree}
\mathbf{p_1} = Dec_{\phi_{1}}(z,\textbf{Attn}(\textbf{H}))
\end{equation}

\begin{equation} \label{eqtwo}
\mathbf{p_i} = Dec_{\phi_{i}}(z,\textbf{Attn}({\mathbf{\{s_{Dec_{\phi_{i-1}}}\}}})) for \; i = 2, ... , n
\end{equation}

where $\mathbf{p_i}$ is a sequence of words in the paraphrase generated by $Dec_{\phi_i}$, \textbf{H} is set of outputs $\{h_{j}^o\}$ generated by the Sentence Encoder $E_{\theta_2}$, $\mathbf{\{s_{Dec_{\phi_{i-1}}}\}}$ are the softmax vectors generated by the previous decoder ($Dec_{\phi_{i-1}}$) and $\textbf{Attn}({\mathbf{\{s_{Dec_{\phi_{i-1}}}\}}})$ are the context vectors obtained by attending over the softmax vectors.

As shown in experimental results section, $Dec_{\phi_i}$ (i$>$1) iteratively improves the output generated by $Dec_{\phi_{i-1}}$. In single decoder model, the output at time-step t is decided based on the outputs at time-steps less than t. In case of multiple decoders, $Dec_{\phi_i}$ (i$>$1) has the information about complete paraphrase generated by $Dec_{\phi_{i-1}}$. We hypothesize that further decoders have prior notion of output to be generated at every time step; this enables them to rectify errors, modify the structure and introduce useful variations.


\subsubsection{Training Technique}
Training objective of our model is similar to the VAE objective function equation \ref{eqone}. To increase the log likelihood of generated paraphrase from all decoders the average of cross entropy (CE) of each decoder output compared to target paraphrase is minimized along with KLD loss. Thus our loss function is:\\
\begin{equation}
\begin{aligned}
\textit{loss}_{VAE-ITERDEC} = \textit{Mean}(CE(Dec_{\phi_{i}})) + \\ D_{KL}(q_{\theta_{1}}(z|x) || q(z))
\end{aligned}
\end{equation}

Also in order to induce variations in the generated paraphrases, we conduct training by sampling three different latent vectors $z_1,z_2,z_3$ and generating the corresponding outputs $o_1,o_2,o_3$. This is done by adding different Gaussian noises to mean and variance vectors obtained corresponding to the input sentence and feeding them to the decoder. We take the final state of the decoder $\{c_i^d\}$ after generating output $\{o_i\}$ as the representation of the corresponding output sentence and minimize pairwise cosine similarity between them by adding the following to the loss function:

\begin{equation}
\textit{loss}_{MultiSample} = \sum_{i,j=1; \; i < j}^{3}\textit{CS}(c_i^d,c_j^d)
\end{equation}

where \textit{CS} denotes cosine similarity. The objective is to tune the model in a way such that different noises added to the mean and variance vector while sampling z results in diverse and different paraphrases while being coherent with the input sentence. We now discuss different experiments conducted for different model variations discussed above.

\section{Experiments} 

\subsection{Datasets}
We present a qualitative and quantitative discussion of the results on two different datasets - Quora question pairs and MSCOCO - across different model variations. Quora dataset\footnote{https://data.quora.com/First-Quora-Dataset-Release-Question-Pairs} comprises of questions asked by the users of the platform and consists of question pairs which are potential paraphrases of each other, as denoted by a binary 1-0 value provided against each pair. We use the pairs with value 1 and discard the remaining ones.  MSCOCO\footnote{http://cocodataset.org/\#download} dataset comprises of about 200k labeled images with each image annotated with 5 captions which are potential paraphrases. We use 2014 release of the dataset which provided separate train and validation splits in order to compare our results with previous baselines and work on paraphrase generation. We randomly select 4 captions out of 5 for each image and randomly divide them into 2 input-paraphrase sentence pairs. Before feeding the sentences to the model for training and inference, we preprocess them by removing punctuations and include only the pairs where both the input sentence and its paraphrase have length $<=15$. Sentences having length less than 15 are padded appropriately using a separate pad token. The number of sentence pairs on which the model is trained and validated after preprocessing is summarized in table \ref{dataset}.

\begin{table}[h]
 \centering
 \caption{Dataset Statistics}
 \label{dataset}
 \begin{tabular}{|c|c|c|}

 \hline 
  Dataset & \# Training Samples & \# Testing Samples \\
  \hline 
  Quora & 87116 & 18773 \\
  \hline 
  MSCOCO & 149438 & 73221 \\
  \hline 
\end{tabular}
\end{table}

\begin{table*}[t]
 \centering
 \caption{METEOR, BLEU and TER scores for different models on test sets of Quora and MSCOCO}
 \label{Results}
 \begin{tabular}{|c|c|c|c|c|c|c|}
 \hline 
  \textbf{Approach} & \multicolumn{3}{|c|}{\textbf{Quora}} & \multicolumn{3}{|c|}{\textbf{MSCOCO}} \\
  \hline
   & METEOR & BLEU & TER & METEOR & BLEU & TER  \\
  \hline
  \textbf{Residual LSTM} \cite{prakash2016neural} & NA & NA& NA&  27.0 & 37.0 & 51.6 \\
  \hline
  \textbf{VAE-SVG} \cite{gupta2017deep} &32.0 & 37.1& 40.8& 30.9& 41.3&40.8 \\
  \hline
  \textbf{VAE-S}& 22.75 & 16.55 & 73.63& 12.11& 4.44& 88.94\\
  \hline
  \textbf{VAE-REF} & 21.85& 14.86& 67.12 & 10.9& 3.36& 84.2\\
  \hline
  \textbf{VAE-VAR} & 25.5& 19.92& 70.01 & 12.38 & 4.7 & 88.73\\
  \hline
  \textbf{VAE-ITERDEC2} & 39.07& 54.19& 32.5& 57.44& 84.64& 7.2\\
  \hline
  \textbf{VAE-ITERVAR} & 39.71& 54.95 & 30.45& \textbf{59.88} & \textbf{87.71} & \textbf{5.8} \\
  \hline
  \textbf{VAE-ITERDEC3} & \textbf{41.95} & \textbf{61.23} & \textbf{26.86}& 53.01& 77.84& 11.85\\
  \hline
\end{tabular}
\end{table*}

\subsection{Implementation Details}
To train our model, we use pre-trained 300 dimensional Glove embeddings\footnote{https://nlp.stanford.edu/projects/glove/} to represent the input words in a sentence and keep them non-trainable. The encoder LSTM in $S_{\theta_1}$ is a single layer LSTM with 600 units. The dimension of mean and variance vectors is kept at 1100 through all the experiments with a batch size of $32$ and learning rate of $5\times 10^{-4}$. $E_{\theta_2}$ and $\{Dec_{\phi_{i}}\}$ are two layer stacked LSTM cells with the number of units in LSTM cell fixed at 600. We have used Adam optimizer \cite{kingma2014adam} for training our model parameters. This configuration is common across different experimental settings.

\subsection{Baseline and Evaluation Measures}
We compare our model with VAE-SVG model \cite{gupta2017deep} which is current state of the art on benchmark datasets and Residual LSTM model \cite{prakash2016neural}. We directly cite the scores as reported in the respective papers. In our work, we do not train the word embedding as done in \cite{gupta2017deep}. To make a fair comparison, we also implemented and trained VAE-SVG model in the same setting. We denote this model as VAE-REF.\\

For quantitative evaluation of our model, we calculate scores on well known evaluations metrics in the domain of machine translation\footnote{We used the software available at https://github.com/jhclark/multeval }: METEOR \cite{lavie2007meteor}, BLEU \cite{papineni2002bleu} and Translation Edit Rate (TER)\cite{snover2006study}. These scores have been shown to correlate well with human judgment. \citeauthor{madnani2012re} \shortcite{madnani2012re} show that these measures perform well for the task of paraphrase recognition also. BLEU score is based on weighted n-gram precision scores of the reference paraphrase with candidate paraphrase. METEOR uses stemming and synonymy detection as well while computing precision and recall. TER measures the edit distance between reference and candidate sentence, so lower the TER better the score.

\subsection{Results}
In order to evaluate our approach, we experimented with following variations of the model: (1) Basic VAE based generative model (VAE-S), (2) VAE-S with reference paraphrase as an additional input (VAE-REF), (3) VAE with attention and $MultiSample$ loss (VAE-VAR), (4) VAE with Iterative decoding having 2 decoders and attention (VAE-ITERDEC2), (5) VAE model comprising of $MultiSample$ loss and 2 decoders with attention (VAE-ITERVAR), and (6) VAE with 3 decoders and attention (VAE-ITERDEC3). Results for each of these models have been summarized in table \ref{Results} for both Quora and MSCOCO datasets. We report all our results and improvements in absolute points.

\begin{figure*}[h]
\minipage{0.5\textwidth}
	\begin{center}
		\includegraphics[width=2.5in,height=2.5in]{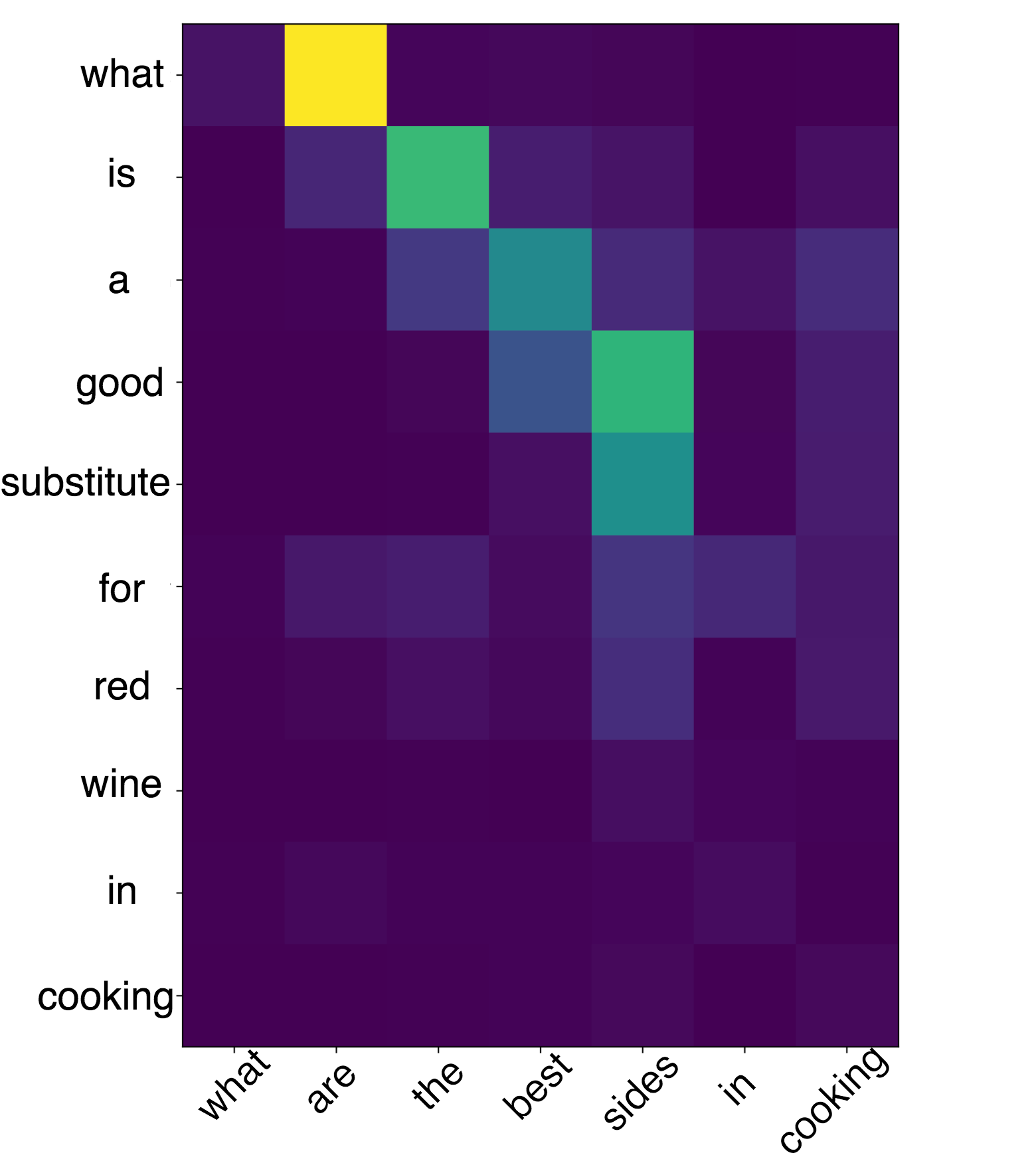}	
	\end{center}
  
\endminipage\hfill
\minipage{0.5\textwidth}
  \includegraphics[width=\linewidth]{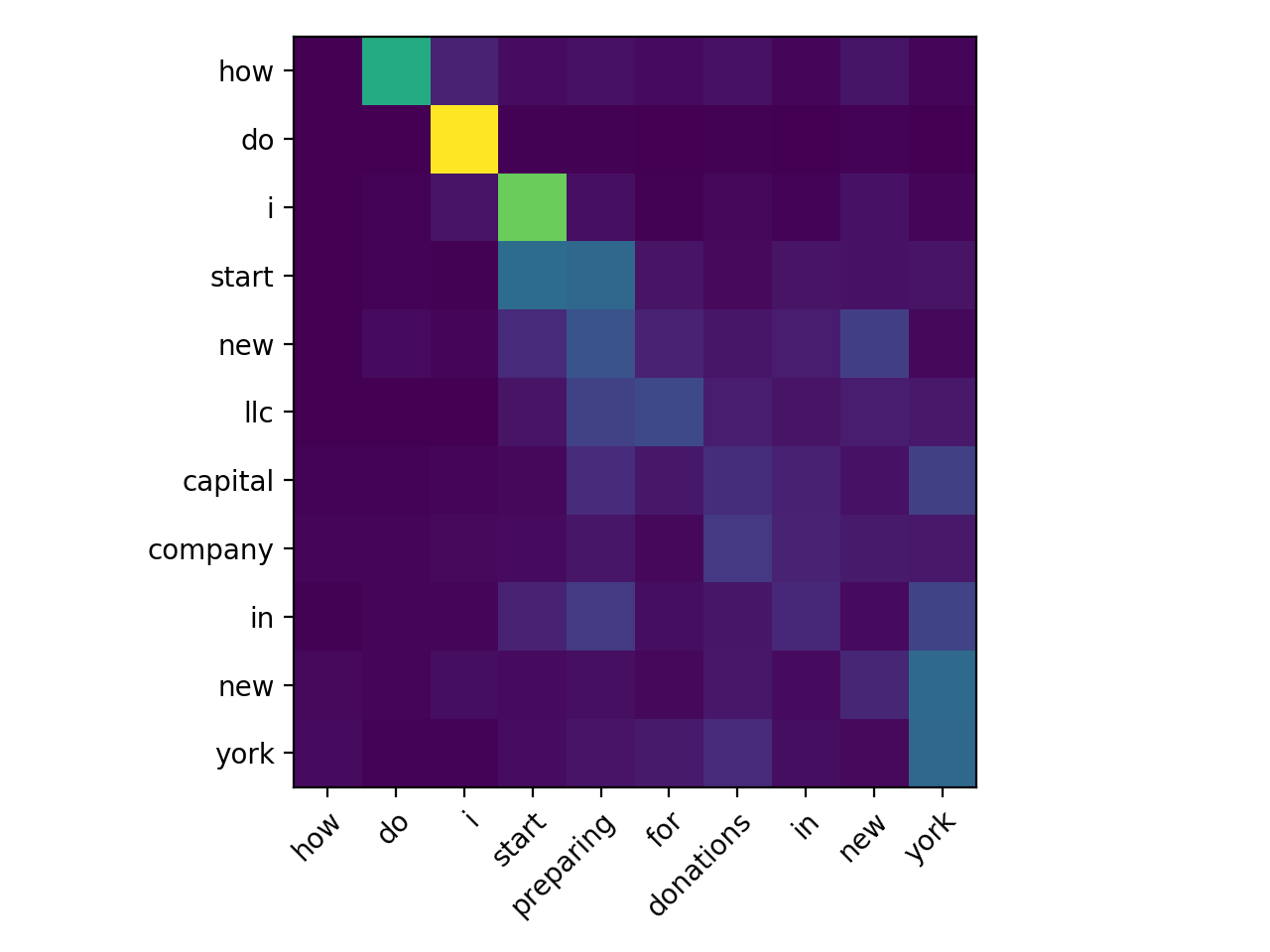}
\endminipage\hfill
\caption{Attention visualization examples from Quora dataset demonstrating parts of paraphrase generated by first decoder (x-axis) on which decoder 2 attends while generating its output (y-axis). Input to the model was - \textit{`what can substitute red wine in cooking'} (left) and \textit{`how do i start an export company or llc in new york city'} (right). It can be seen that second decoder corrects the output of the first one by attending on incorrect words while replacing them with a better phrase.}
\label{quora_heat_maps}
\end{figure*}

\begin{table}[h]
 \centering
 \caption{Few examples of paraphrases generated by our VAE-ITERDEC2 model on Quora dataset}
 \label{Quora_examples}
 \begin{tabular}{|l|l|l|l|}
 \hline 
  \textbf{Input} & what are the top universities for computer   \\ & science in the world \\
  \hline 
  \textbf{Decoder 1} & what are the \textcolor{red}{best universities} for computer  \\ & science in the world\\
  \hline
  \textbf{Decoder 2} & what are the \textcolor{ForestGreen}{best computer science} colleges\\
  \hline
  \textbf{Expected} & what are the \textcolor{blue}{best computer science} schools\\
  \hline
  \hline
  \textbf{Input} & which is best time for exercise\\
  \hline 
  \textbf{Decoder 1} & what is best time exercise \\
  \hline
  \textbf{Decoder 2} & \textcolor{ForestGreen}{when} is the best time \textcolor{ForestGreen}{to exercise}\\
  \hline
  \textbf{Expected} & \textcolor{blue}{when} is the best time \textcolor{blue}{to workout}\\
  \hline
  \hline
  \textbf{Input} & what can substitute red wine in cooking\\
  \hline 
  \textbf{Decoder 1} & what are the \textcolor{red}{best sides} in cooking \\
  \hline
  \textbf{Decoder 2} & what is a \textcolor{ForestGreen}{good substitute for red wine} in \\ & cooking\\
  \hline
  \textbf{Expected} & what is a \textcolor{blue}{good replacement for red wine} in \\ & cooking\\
  \hline
  \hline
  \textbf{Input} & how do i start an export company or llc in \\ & new york city\\
  \hline 
  \textbf{Decoder 1} & how do i start \textcolor{red}{preparing for donations} in \\ &new york \\
  \hline
  \textbf{Decoder 2} & how do i start \textcolor{ForestGreen}{new llc capital company} in \\ &new york\\
  \hline
  \textbf{Expected} & how do i start an import/export \textcolor{blue}{llc} in new \\ & york city\\
  \hline
\end{tabular}
\end{table}

\subsubsection{Quora} As we can see our proposed iterative decoding mechanism improves the score by a huge margin as compared to baseline VAE-S. The improved scores are better than any other previous work done in paraphrase generation - with near $7\%$ and $17\%$ absolute increase in $METEOR$ and $BLEU$ scores respectively compared to the previous best scores \cite{gupta2017deep} - thus establishing a new state of the art in this task. Our $TER$ score is also $8.3\%$ less than the previously established best score. Table \ref{Quora_examples} shows a comparison between paraphrases generated by the first decoder and the improvements made by the second decoder on a few example sentences.


In some cases, as in first example in table \ref{Quora_examples}, the output generated by the second decoder resembles the expected paraphrase more than the paraphrase generated by the first decoder which leads to a better score - the first decoder just replaces the word \textit{`top'} in the input sentence with \textit{`best'} while the second decoder changes the sentence structure by introducing the phrase \textit{`best computer science'} which also matches with the expected paraphrase. Another observation is that many times the second decoder makes the generated paraphrase correct and semantically more similar to the input sentence than the output of the first decoder like in the third and last example in table \ref{Quora_examples}. Figure \ref{quora_heat_maps} shows attention heatmaps demonstrating the phrases in the output of the first decoder where the second decoder focuses while generating the paraphrase. For the last example in table \ref{Quora_examples} it can be seen in figure \ref{quora_heat_maps} (right) that the second decoder attends on \textit{`start preparing for donations'} while replacing it with \textit{`start new llc'}. Similarly for the third example in table \ref{Quora_examples} the second decoder generates \textit{`good substitute'} while attending on \textit{`best sides'} - as can be seen in figure \ref{quora_heat_maps} (left). Thus the second decoder is focusing on mistakes in the previous output to make a guided decision while generating output.

On adding the $MultiSample$ loss to VAE-ITERDEC2 model, $TER$ reduces by $2\%$. We also extended the VAE-ITERDEC2 model (without $MultiSample$ loss) by using an additional decoder resulting in 3 decoders which further boosted up the $METEOR$ score to $41.95\%$, $BLEU$ to $61.23\%$ and reduced $TER$ to $26.86\%$.  

\begin{table}[h!]
 \centering
 \caption{Few examples of paraphrases generated by our VAE-ITERDEC2 model on MSCOCO dataset}
 \label{MSCOCO_examples}
 \begin{tabular}{|l|l|l|l|}
 \hline 
  \textbf{Input} & a group of motorcyclists are driving down \\ & the city street \\
  \hline 
  \textbf{Decoder 1} & a group of \textcolor{red}{people that are sitting} on a street \\
  \hline
  \textbf{Decoder 2} & a group of \textcolor{ForestGreen}{motorcycles drive down} a city \\ & street\\
  \hline
  \textbf{Expected} & a group of \textcolor{blue}{motorcycles drive down} a city \\ & street\\
  \hline
  \hline
  \textbf{Input} & a man sits with a traditionally decorated \\ & cow\\
  \hline 
  \textbf{Decoder 1} & a man is sitting \textcolor{red}{on a large grill in a} \\ & \textcolor{red}{restaurant}\\
  \hline
  \textbf{Decoder 2} & an \textcolor{ForestGreen}{equestrian} man in \textcolor{ForestGreen}{armor costume} sitting \\ & with a decorated cow\\
  \hline
  \textbf{Expected} & an \textcolor{blue}{indian} man in \textcolor{blue}{religious attire} sitting with \\ & a decorated cow\\
  \hline
  \hline
  \textbf{Input} & a beautiful dessert waiting to be shared by \\ & two people\\
  \hline 
  \textbf{Decoder 1} & a table with \textcolor{red}{three plates of food} and a fork \\
  \hline
  \textbf{Decoder 2} &there is a \textcolor{ForestGreen}{piece of cake} on a plate \textcolor{ForestGreen}{with} \\ & \textcolor{ForestGreen}{flowers} on it\\
  \hline
  \textbf{Expected} & there is a \textcolor{blue}{piece of cake} on a plate \textcolor{blue}{with} \\ & \textcolor{blue}{decorations} on it\\
  \hline
  \hline
  \textbf{Input} & a home office with laptop printer scanner \\ & and extra monitor\\
  \hline 
  \textbf{Decoder 1} & a desk with a laptop and \textcolor{red}{a mouse}\\
  \hline
  \textbf{Decoder 2} & \textcolor{ForestGreen}{office setting} with \textcolor{ForestGreen}{office equipment} on \textcolor{ForestGreen}{desk} \\ & \textcolor{ForestGreen}{top}\\
  \hline
  \textbf{Expected} & \textcolor{blue}{office space} with \textcolor{blue}{office equipment} on \textcolor{blue}{desk} \\ & \textcolor{blue}{top}\\
  \hline
\end{tabular}
\end{table}

\begin{figure*}[h]
\minipage{0.5\textwidth}
	\begin{center}
		\includegraphics[width=2.5in,height=2.2in]{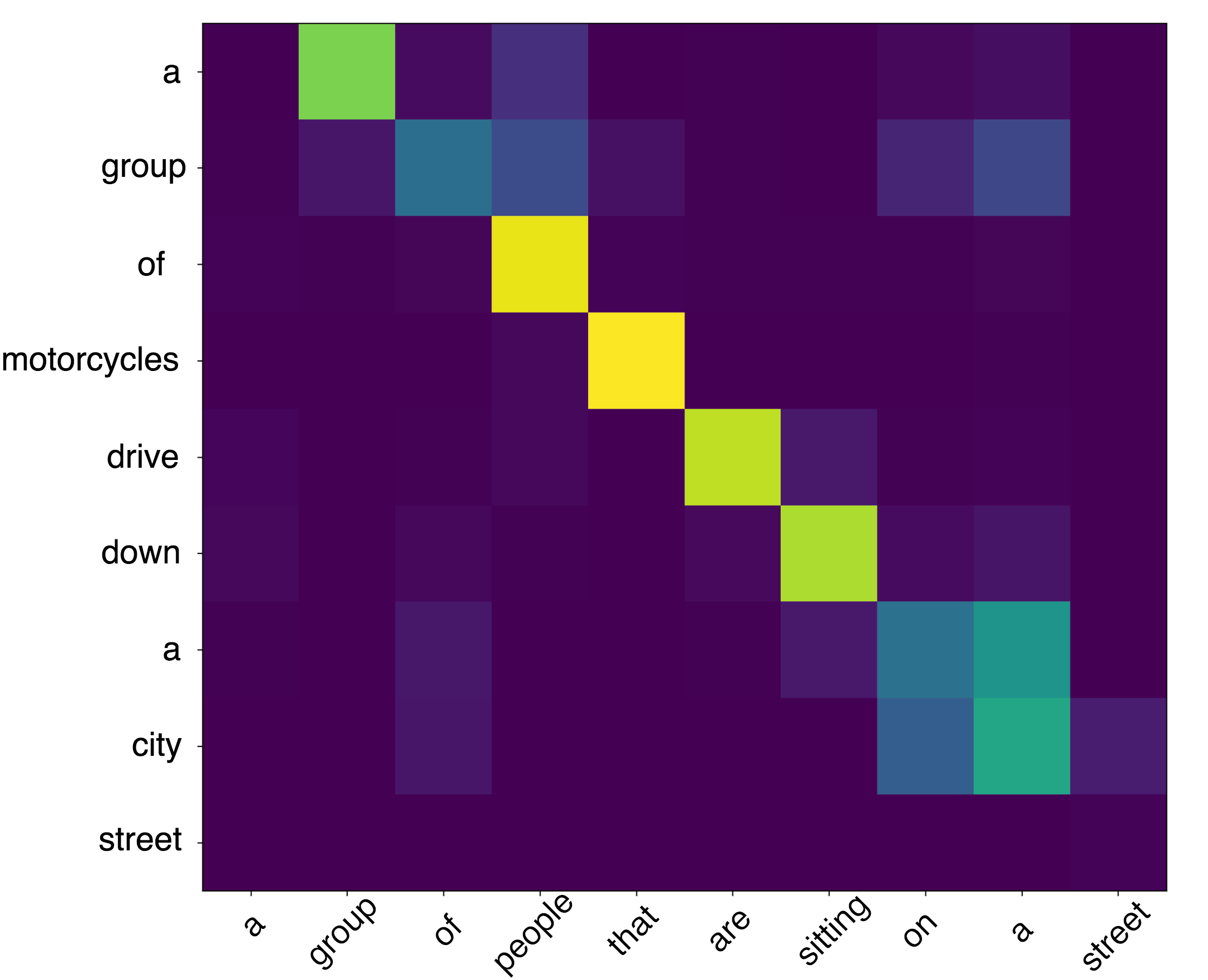}	
	\end{center}
  
\endminipage\hfill
\minipage{0.5\textwidth}
  \includegraphics[width=2.5in,height=2.5in]{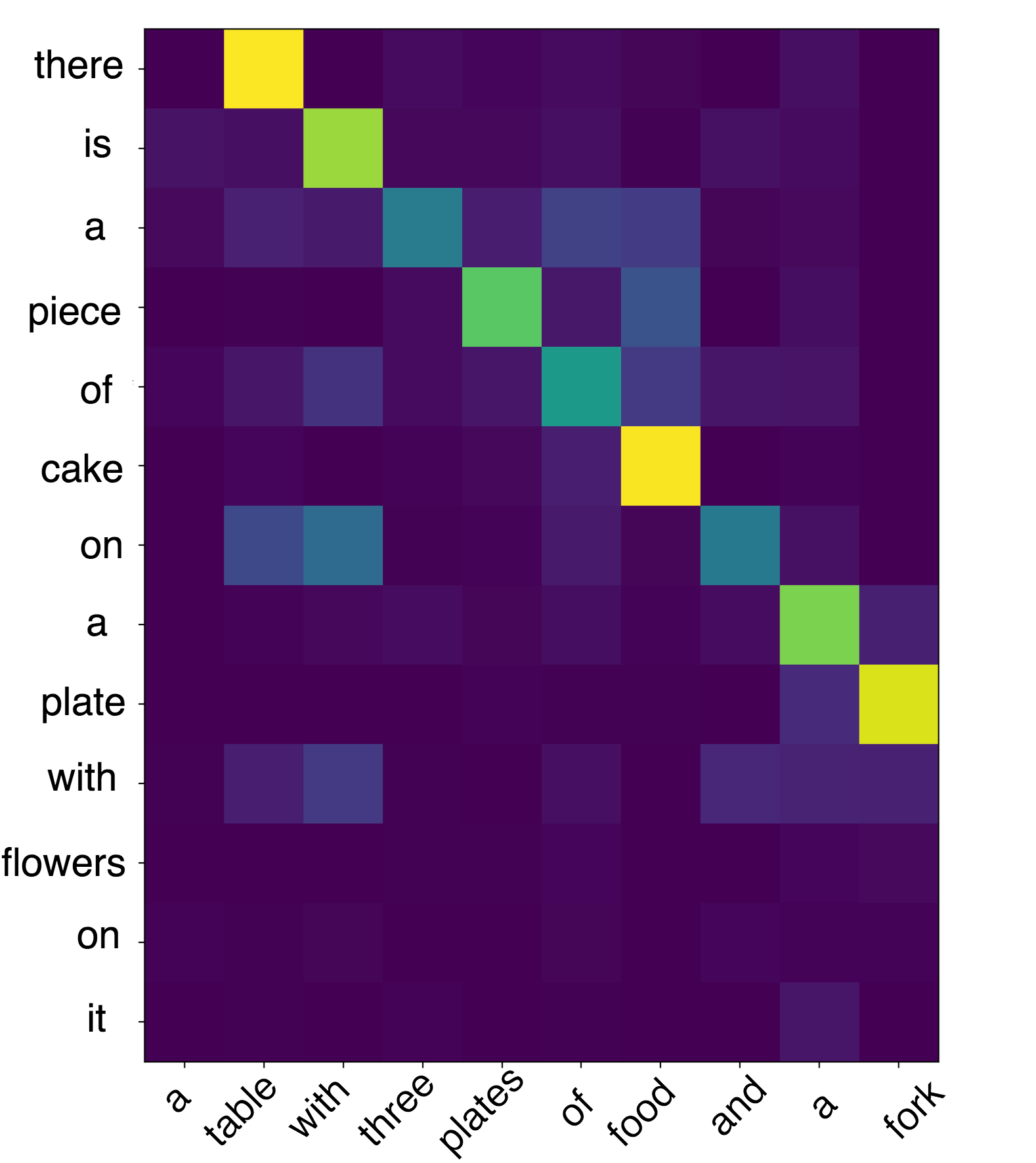}
\endminipage\hfill
\caption{Attention visualization examples from MSCOCO dataset demonstrating parts of paraphrase generated by first decoder (x-axis) on which decoder 2 attends while generating its output (y-axis). Input to the model was - \textit{`a group of motorcyclists are driving down the city street'} (left) and \textit{`a beautiful dessert waiting to be shared by two people'} (right).}
\label{coco_heat_maps}
\end{figure*}

\begin{table*}[t]
 \centering
 \caption{Comparison of METEOR, BLEU and TER scores of output of decoder 1 with - expected paraphrase and output of decoder 2 - in VAE-ITERDEC2 model on test sets of Quora and MSCOCO}
 \label{Results_dec1}
 \begin{tabular}{|c|c|c|c|c|c|c|}
 \hline 
  \textbf{Comparison of decoder 1 output} & \multicolumn{3}{|c|}{\textbf{Quora}} & \multicolumn{3}{|c|}{\textbf{MSCOCO}} \\
  \hline
   & METEOR & BLEU & TER & METEOR & BLEU & TER  \\
  \hline
  \textbf{with expected paraphrase} & 27.09 & 22.12 & 67.12 & 15.15 & 8.09 & 79.52 \\
  \hline
  \textbf{with decoder 2 output} & 26.2 & 23.19 & 68.52 & 14.99 & 8.02 & 79.65 \\
  \hline
\end{tabular}
\end{table*}

\subsubsection{MSCOCO} Our VAE-ITERDEC2 model provides significant improvements on this dataset outperforming previously best approaches on all three metrics with $57.44\%$ $METEOR$ score, $84.64\%$ $BLEU$ and $7.2\%$ $TER$. This is an improvement of over $26.5\%$, $43.3\%$ and reduction of $33.6\%$ in $METEOR$, $BLEU$ and $TER$ respectively compared to the previous state of the art. Contrary to Quora, however, VAE-ITERDEC3 attains slightly less score compared to VAE-ITERDEC2 in terms of these metrics which shows that addition of the third decoder does not necessarily lead to better results. But using the second decoder significantly improves the results. Thus it still needs to be explored what is the optimal number of decoders needed for a dataset or if it can be decided dynamically. Adding $MultiSample$ loss to VAE-ITERDEC2 gives best results giving a $METEOR$ score of $59.88\%$, $BLEU \; 87.71\%$ and $TER \; 5.8\%$. Few example paraphrases generated by VAE-ITERDEC2 on MSCOCO have been shown in table \ref{MSCOCO_examples}.

In the first example, first decoder generates a paraphrase which has little relevance with respect to the input, however, the second decoder corrects it by replacing \textit{`group of people that are sitting'} with \textit{`group of motorcycles drive down'} as can be seen in the attention map also in figure \ref{coco_heat_maps} (left). In the third example in table \ref{MSCOCO_examples} first decoder uses a generic term \textit{`food'} as a replacement for \textit{`desert'} while the second decoder introduces the word \textit{`cake'} while attending on \textit{`food'} as can be seen in the attention visualization in figure \ref{coco_heat_maps} (right). It also introduces \textit{`with flowers on it'} to represent the notion of \textit{`beautiful dessert'} in the original sentence. Similarly in the last example in the table, the paraphrase generated by the second decoder includes \textit{`office setting'}, making it coherent with the input while its structure resembles the expected paraphrase.

To compare the outputs generated by the two decoders in VAE-ITERDEC2 model, we computed the metric scores of decoder 1 output with - expected paraphrase and decoder 2 output as shown in table \ref{Results_dec1}. $METEOR$ score with expected paraphrase is sufficiently low compared to VAE-ITERDEC2 scores in table \ref{Results}. This implies that the second decoder significantly improves the $METEOR$ scores over the first decoder. The same observation holds for $BLEU$ and $TER$. Comparing decoder 1 output with decoder 2 outputs, we get a high $TER$ which suggests second decoder generates sufficiently different outputs from the first one. 

\section{Conclusion}
In this paper, we have proposed attention based ReDecode framework for iterative refinement of generated paraphrases using VAE based Seq2Seq model. It comprises of a sequence of decoders which generate paraphrases turn by turn. Given a decoder, it attends on the output generated by the preceding decoder and modifies it by rectifying errors and introducing semantically coherent phrases, while generating its output. Quantitatively, it improves the previous best scores on standard metrics and benchmark datasets, establishing a new state of the art in this task. 

We experimented with maximum three decoders using our ReDecode framework. On Quora dataset, using three decoders improved the scores over two decoders model contrary to MSCOCO. Determining the optimal number of decoders, which can be dataset dependent, remains future work. Furthermore, the proposed architecture is generic and might be beneficial in other sequence generation tasks such as machine translation.

\bibliographystyle{aaai}
\bibliography{references}

\end{document}